%% file: root.tex
\documentclass[letterpaper, 10 pt, conference]{ieeeconf}  % Comment this line out if you need a4paper

\IEEEoverridecommandlockouts                              % This command is only needed if 
\usepackage{amsmath,amsfonts}
\usepackage{amssymb}

\usepackage{algorithm}
\usepackage[noend]{algorithmic}
\usepackage{color}

\usepackage{array}
\usepackage{textcomp}
\usepackage{stfloats}

\usepackage{verbatim}
\usepackage{graphicx}

\usepackage{cite}
\usepackage[utf8]{inputenc} % allow utf-8 input
\usepackage[T1]{fontenc}    % use 8-bit T1 fonts
\usepackage{hyperref}       % hyperlinks
\usepackage{url}            % simple URL typesetting
\usepackage{booktabs}       % professional-quality tables
\usepackage{amsfonts}       % blackboard math symbols
\usepackage{nicefrac}       % compact symbols for 1/2, etc.
\usepackage{microtype}      % microtypography
\usepackage{lipsum}
\usepackage{fancyhdr}       % header
\usepackage{graphicx}       % graphics
\usepackage{wrapfig}
\usepackage{subfigure}
\usepackage{multirow}
\usepackage{threeparttable}
\usepackage{makecell}

\usepackage{pifont}
\newcommand{\xmark}{\ding{53}}%

\title{\LARGE \bf
Generalizable Articulated Object Perception with Superpoints
}

\author{Qiaojun Yu$^{1*}$, Ce Hao$^{2*}$, Xibin Yuan$^{1*}$, Li Zhang$^{3}$, Liu Liu$^{4}$, Yukang Huo$^{5}$, Rohit Agarwal$^{6}$ and Cewu Lu$^{1+}$% <-this % stops a space
\thanks{$^{*}$ indicates the equal contribution. $^{+}$ indicates the corresponding author.}%
\thanks{$^{1}$Qiaojun Yu, Xibin Yuan and Cewu Lu are with Department of Computer Science, Shanghai Jiao Tong University, China. {\tt\small \{yqjllxs,  2022yxb, lucewu\}@sjtu.edu.cn}}%
\thanks{$^{2}$Ce Hao is with Department of Computer Science, National University of Singapore, Singapore.
        {\tt\small cehao@u.nus.edu}}%
\thanks{$^{3}$Li Zhang is with University of Science and Technology of China, China. 
        {\tt\small zanly20@mail.ustc.edu.cn}}%
\thanks{$^4$Liu Liu is with Department of Computer Science and Information Engineering, Hefei University of Technology, China, \texttt{liuliu@hfut.edu.cn}}%
\thanks{$^{5}$Yukang Huo is with China Agricultural University, China. 
        {\tt\small huoyukang@cau.edu.cn}}%
\thanks{$^{6}$Rohit Agarwal is Department of Computer Science and Engineering, National Institute of Technology, Durgapur, India. {\tt\small ra.22cs1102@phd.nitdgp.ac.in}}%  
}

\begin{document}

\maketitle
\thispagestyle{empty}
\pagestyle{empty}

%%%%%%%%%%%%%%%%%%%%%%%%%%%%%%%%%%%%%%%%%%%%%%%%%%%%%%%%%%%%%%%%%%%%%%%%%%%%%%%%

\input{sections/0_Abstract}
\input{sections/1_Introduction}
\input{sections/3_Preliminary}
\input{sections/4_Method}
\input{sections/5_Experiment}
\input{sections/6_Conclusion}

\clearpage
%%%%%%%%%%%%%%%%%%%%%%%%%%%%%%%%%%%%%%%%%%%%%%%%%%%%%%%%%%%%%%%%%%%%%%%%%%%%%%%%
\bibliographystyle{IEEEtran}
\bibliography{references}

\end{document}

%% file: sections/0_Abstract.tex
\begin{abstract}
Manipulating articulated objects with robotic arms is challenging due to the complex kinematic structure, which requires precise part segmentation for efficient manipulation. In this work, we introduce a novel superpoint-based perception method designed to improve part segmentation in 3D point clouds of articulated objects. We propose a learnable, part-aware superpoint generation technique that efficiently groups points based on their geometric and semantic similarities, resulting in clearer part boundaries. Furthermore, by leveraging the segmentation capabilities of the 2D foundation model SAM, we identify the centers of pixel regions and select corresponding superpoints as candidate query points. Integrating a query-based transformer decoder further enhances our method's ability to achieve precise part segmentation. Experimental results on the GAPartNet dataset show that our method outperforms existing state-of-the-art approaches in cross-category part segmentation, achieving AP50 scores of $77.9\%$ for seen categories ($4.4\%$  improvement) and $39.3\%$ for unseen categories ($11.6\%$ improvement), with superior results in $5$ out of $9$ part categories for seen objects and outperforming all previous methods across all part categories for unseen objects.

\end{abstract}
% \begin{IEEEkeywords}
% Articulated Objects, Pose Estimation
% \end{IEEEkeywords}

%% file: sections/1_Introduction.tex
\section{Introduction} \label{Sec: intro}
Articulated objects, such as doors and drawers, are ubiquitous in daily life due to their kinematic connections. As embodied intelligence technology continues to advance, it becomes increasingly important for robots to not only recognize these objects~\cite{zhang2024vocapter,liu2024kpa, yu2024manipose, geng2023gapartnet} but also manipulate them effectively by performing tasks like opening doors, closing drawers, or even lifting pot lids. 
Existing approaches based on reinforcement learning (RL) and imitation learning typically address the manipulation of articulated objects by predicting affordances and generating motion trajectories through learned policies~\cite{geng2022end, mo2021where2act, liu2023category, wu2021vat, ning2024where2explore}. However, these methods often face significant challenges in generalizing to unseen objects, particularly when variations in object geometry are introduced, thereby limiting the transferability of the learned skills.

In contrast, by leveraging powerful vision models, part segmentation-based approaches to articulated object modeling offer a more general solution, achieving accurate perception of articulated objects, forming  the foundation for successful manipulation~\cite{liu2024survey, weng2021captra, jain2021screwnet}. This precise and efficient perception enables robots to handle complex tasks with greater reliability~\cite{Flowbot++, jiang2022ditto, yu2023gamma}. However, while these approaches provide significant advantages, prior part segmentation methods typically segment 3D point clouds into different parts based on individual point clouds~\cite{liu2023category, qi2017pointnet++, li2020category, zhangrethinking}. Although these methods can achieve high modeling accuracy with familiar objects, they struggle to extract transferable information in the face of complex variations in point clouds, significantly reducing segmentation accuracy with unseen objects.  
Superpoint-based methods~\cite{robert2023efficient, hui2021superpoint, kolodiazhnyi2024oneformer3d} partition point clouds into point sets, known as superpoints, which are groups of neighboring points adapted to local complexity and aggregating geometric information. This superpoint-based approach not only reduces computational overhead but also enhances the model's ability to generalize across diverse object geometries by effectively integrating local geometric features, thereby improving segmentation accuracy with unseen objects.

In this paper, we introduce \textbf{G}eneralizable \textbf{A}rticulated Object \textbf{P}erception with \textbf{S}uperpoints (\textbf{GAPS}), a novel approach designed to enhance part segmentation in diverse articulated objects within 3D point clouds.GAPS improves superpoint boundary clarity through learnable part-aware superpoint generation techniques, ensuring more distinct superpoints. Building on this, it leverages the 2D foundation model SAM~\cite{kirillov2023segment} to effectively segment pixel regions, where each region’s center uniquely identifies the corresponding 3D superpoints. These superpoints are then used as query points for part segmentation, enabling a more generalizable and adaptable selection of query points.  By utilizing a query-based transformer decoder, GAPS achieves precise part segmentation across articulated objects. The main contributions of this paper are as follows:

1) We design the learnable part-aware superpoint generation method that groups point clouds as superpoints based on geometric and semantic similarities. Compared to rule-based superpoint generation, our approach is more effective in handling smaller parts and achieving clearer boundaries.

% 2) By leveraging the 2D foundation model SAM's ability to segment images into pixel regions, where each region's center uniquely corresponds to a 3D superpoint, these key 3D superpoints serve as candidates for 3D query points. This allows the query-based transformer decoder to capture and utilize local information within each part more effectively, enabling GAPS to achieve precise part segmentation across cross-category articulated objects.
2) The 2D foundation model SAM segments images into pixel regions, with each center mapping to a unique 3D superpoint. These superpoints act as query points, allowing the transformer decoder to effectively capture local information, enabling GAPS to achieve precise part segmentation across diverse articulated objects.

3) We conduct experiments on the articulated object modeling benchmark GAPartNet~\cite{geng2023gapartnet}, where GAPS outperforms existing state-of-the-art part segmentation methods in both seen objects and unseen cross-category generalization.

%% file: sections/3_Preliminary.tex
\section{Preliminary} \label{Sec: prelim}

We formulate the articulated object part segmentation task $\mathcal{T}$ as follows. An articulated object $\mathcal{M}$ consists of $K$ variable movable parts, represented as $\mathcal{M}=\{m_{i}\}_{i=1}^{K}$. We observe the object $\mathcal{M}$ using RGB-D cameras and project it into a point cloud $P$ with $N$ points,  $P=\{\mathbf{p}_{i}\in \mathbb{R}^{3}\}_{i=1}^{N}$. 
Superpoints are an over-segmented set of point clouds that adapt to local geometric structures and capture contextual features. Given a point cloud as $P$ with $N$ points as $P=\{\mathbf{p}_{i}\in \mathbb{R}^{3}\}_{i=1}^{N}$ and its corresponding features $F=\{\mathbf{f}_{i}\in \mathbb{R}^{d}\}_{i=1}^{N}$, superpoint generation aims to construct $O$ superpoints as $S=\{\mathbf{s}_{i}\in \mathbb{R}^{3}\}_{i=1}^{O}$ and its corresponding features $E=\{\mathbf{e}_{i}\in \mathbb{R}^{d}\}_{i=1}^{o}$ from the point cloud $P$, assigning each point to one of the 
$O$ superpoint centers with the highest probability. In this way, the superpoints $S$ with corresponding features $E$ can be used to represent the entire point cloud, with each superpoint encoding both local geometric and semantic features.
% \textcolor{red}{If teaser has the illustration, it is fine. Otherwise, we should have a figure to illustrate superpoints.}

%% file: sections/4_Method.tex
\section{Method} \label{Sec: method}

\begin{figure*}
    \centering
    \includegraphics[width=0.8\textwidth]{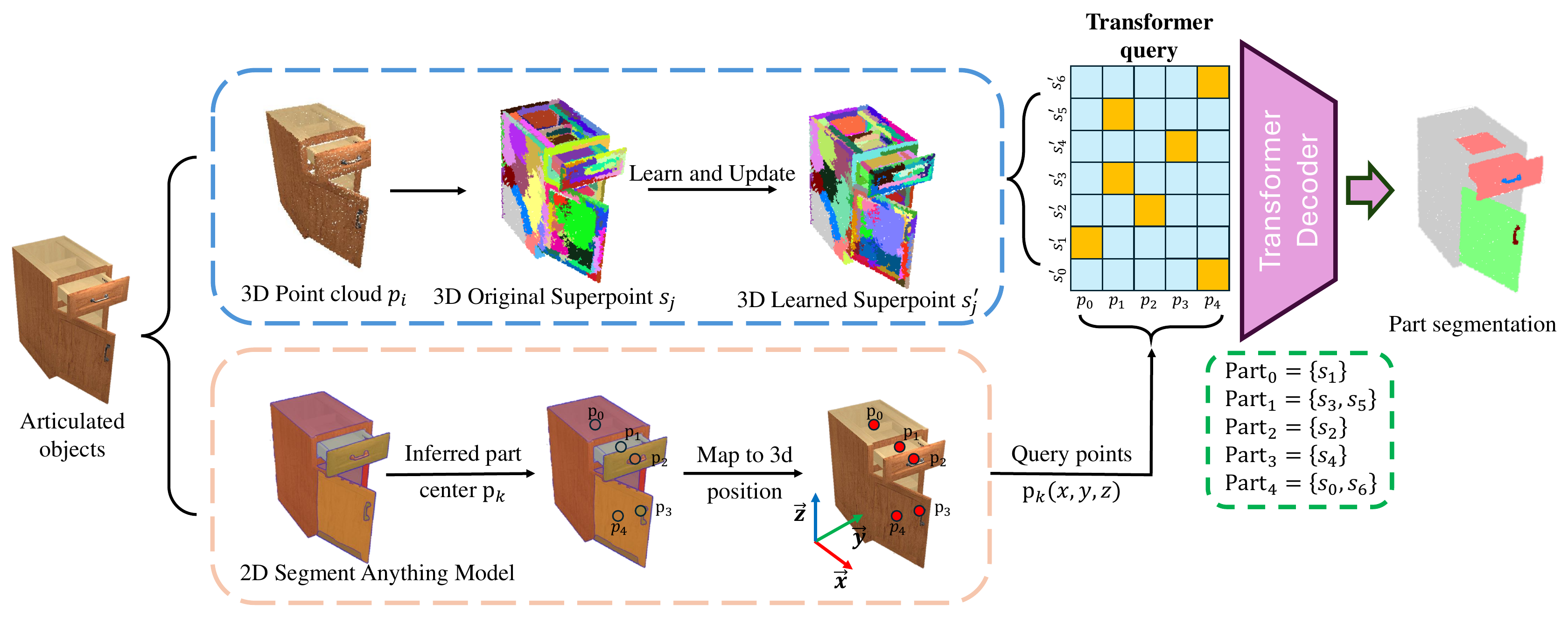}
    \vspace{-1mm}
    \caption{GAPS segments articulated objects into semantic parts. It leverages both 3D point clouds to cluster superpoints and 2D image segmentation to infer part center, queried by a transformer decoder for part segmentation.}
    \label{Fig: pipeline}
\vspace{-4mm}
\end{figure*}

In this section, we present an innovative methodology GAPS, as shown in Fig~\ref{Fig: pipeline}. For a given single-view point cloud of an object, we leverage the Point Transformer-V2~\cite{wu2022point} to extract point-wise features, which are then processed through a part-aware superpoint generation module to produce superpoints, resulting in sharper and more well-defined boundaries between superpoints (Section~\ref{Subsec: sp gen}). By utilizing the SAM-guided 2D information to identify corresponding 3D superpoints as candidate 3D query points and combining it with a transformer decoder, we enable accurate part segmentation of articulated objects (Section~\ref{Subsec: sp seg}).

\subsection{Part-aware Superpoint Generation} \label{Subsec: sp gen}

Superpoints are groups of 3D points semantically clustered based on similar geometric features. In articulated object part segmentation, we leverage these superpoints to capture local geometric information, enhancing the model's generalization ability and improving segmentation accuracy. Unlike previous methods focused on instance segmentation, our task deals with the complexity of articulated part segmentation, where parts vary significantly in size and have intricate connections. To address this, we draw inspiration from SPNet~\cite{hui2021superpoint} and employ a learnable soft association map to model relationships between points and superpoints. This approach generates part-aware superpoints, effectively addressing the challenges of articulated part segmentation.

Given an articulated object represented by a point cloud $P=\{\mathbf{p}_{i}\in \mathbb{R}^{3}\}_{i=1}^{N}$, we initially use hand-crafted features~\cite{guinard2017weakly} to construct a hard point-superpoint assignment. The superpoint coordinates \(S\) and features \(E\) are derived as the averages of the coordinates and features of the points assigned to each superpoint. However, this initial over-segmentation, achieved through unsupervised optimization, may not effectively capture fine part instances, leading to issues such as cross-part and nested-part segmentation.
To address this, we apply a refinement process that updates both the point-superpoint assignment and the soft association map, improving segmentation accuracy by better handling these complexities. To further enhance computational efficiency, we selectively build the association map using only the nearest $6$ superpoints to each point, denoted as \(A \in \mathbb{R}^{N \times 6}\). Specifically, a cosine-similarity-like operation, implemented through an MLP, is used to update the point-superpoint association map. The association between the \(i\)-th point and \(j\)-th superpoint $a_{ij}$, is updated as follows:

\begin{equation}
a_{ij} = \phi(p_i, s_j)g(p_i) \cdot \psi(f_i, e_j)h(f_i),
\end{equation}

where $\phi(\cdot, \cdot): \mathbb{R}^{3} \times \mathbb{R}^{3} \rightarrow \mathbb{R}^{c}$ and $g(\cdot): \mathbb{R}^{3} \rightarrow \mathbb{R}^{c}$ are two mapping functions in the coordinate space, while $\psi(\cdot, \cdot): \mathbb{R}^{d} \times \mathbb{R}^{d} \rightarrow \mathbb{R}^{c}$ and $h(\cdot): \mathbb{R}^{d} \rightarrow \mathbb{R}^{c}$ are two mapping functions in the feature space, all implemented by MLP. Then we normalize the association map of each point:
\begin{equation}
\tilde{a}_{ij} = \frac{\operatorname{exp}(a_{ij})}{\sum_{k=1}^6 \operatorname{exp}(a_{ik})}.
\end{equation}

After updating the association map, we use the normalized association scores as weights to update the superpoints' coordinates and features as follows:
\begin{equation}
s_j = \frac{\sum^N_{i=1} \tilde{a}_{ij} \cdot p_i}{\sum^N_{i=1} \tilde{a}_{ij}}, e_j = \frac{\sum^N_{i=1} \tilde{a}_{ij} \cdot f_i}{\sum^N_{i=1} \tilde{a}_{ij}}.
\end{equation}

We assume that points within the same superpoint belong to the same object part, and we use one-hot encoded labels corresponding to the parts. Given the one-hot labels \( L = \{l_i \in \mathbb{R}^K\}_{i=1}^N \) for each point, the labels for each superpoint can be computed through a weighted average as follows: $L^s=\{l^s_j = \frac{\sum^N_{i=1} \tilde{a}_{ij} \cdot l_i}{\sum^N_{i=1} \tilde{a}_{ij}}\}_{j=1}^O$. We then reconstruct the point labels \(\tilde{L} = \{\tilde{l}_i = \sum_{j=1}^M \tilde{a}_{ij} \cdot l^s_j\}_{i=1}^N\) using normalized association scores. Additionally, pseudo labels \(\tilde{L}^s = \{\tilde{l}^s_j = \operatorname{mod}(\sum_{i=1}^N a_{ij} \cdot l_i)\}_{j=1}^M\) can be derived for the superpoints using a voting mechanism based on point-superpoint correlations. The corresponding loss is then defined as follows:

\begin{equation}
\mathcal{L}_{sp} = \frac{1}{N} \sum^N_{i=1} \mathcal{L}(l_i, \tilde{l}_i) + \frac{1}{O} \sum^O_{j=1} \mathcal{L}(l^s_j, \tilde{l}^s_j),
\end{equation}

where $\mathcal{L}(\cdot, \cdot)$ is the cross-entropy loss function. This loss encourages part-aware consistency and ensures the superpoint structure aligns with part boundaries.

\subsection{Superpoint-based Part Segmentation} \label{Subsec: sp seg}
% Our methodology for part segmentation of articulated objects leverages superpoints, enhanced by the integration of SAM-guided 2D information to identify corresponding 3D superpoints as candidate 3D query points. These superpoints, constructed from single-view point clouds in the part-aware superpoint generation stage, capture local geometric properties, resulting in rich feature encoding that is sensitive to the nuances of articulated parts and robust against variations in scale and size. Building upon the SPFormer framework~\cite{sun2023superpoint}, we employ a 6-layer query decoder to process the features extracted from the superpoints, further refining the segmentation process.

Our approach to part segmentation of articulated objects leverages superpoints, enhanced by integrating SAM-guided 2D information to identify corresponding 3D superpoints as candidate query points. These superpoints, generated from single-view point clouds during the part-aware superpoint generation stage, capture local geometric features, providing rich encoding sensitive to the nuances of articulated parts and robust to scale and size variations. Building on the SPFormer framework~\cite{sun2023superpoint}, we employ a 6-layer query decoder to refine the segmentation process further.

\textbf{Query Decoder Architecture.}
Each superpoint, represented by the aggregated coordinates and features of its constituent points, forms the basis of our segmentation approach. The coordinates of these SAM-selected 3D query points are used to generate position embeddings, which serve as queries. The superpoint features, combined with their corresponding position embeddings, are then fed into a query decoder, employing a 6-layer transformer decoder architecture. This structure leverages cross-attention mechanisms, where each superpoint query attends over all points to refine its representation.  Let $S = \{s_1, s_2, ..., s_O\}$ be the set of superpoint feature vectors, where $s_j \in \mathbb{R}^d$ is the feature vector for the $j$-th superpoint, and $O$ is the total number of superpoints. The corss-attention operation for the $l$-th layer of the decoder can be defined as:
\begin{equation}
\text{CrossAttention}^{(l)}(Q, K, V) = \text{softmax}\left(\frac{QK^T}{\sqrt{d_k}}\right)V
\end{equation}
where $Q$, $K$, and $V$ are the query, key, and value matrices derived from the superpoint features, respectively. The operation computes the attention-weighted sum over the values $V$, where the attention weights are determined by the compatibility between the queries $Q$ and keys $K$. The scaling factor $\frac{1}{\sqrt{d_k}}$ ensures stability during training.

The output of the cross-attention layers is then passed through feed-forward networks within each layer of the decoder to progressively refine the superpoint representations. The final superpoint query representation at the $l$-th layer is given by:
\begin{equation}
s_j^{(l)} = \text{FFN}\left(\text{CrossAttention}^{(l)}(Q^{(l)}, K^{(l)}, V^{(l)})\right)
\end{equation}
where $\text{FFN}$ denotes a feed-forward network that further processes the output of the cross-attention mechanism.
% By stacking multiple layers of this architecture, the model is able to capture complex relationships and refine the superpoint queries in a hierarchical manner, ultimately leading to enhanced part segmentation capabilities.
%
%
% The extracted superpoint features are fed into a query decoder that employs a 6-layer structure. 
%
Due to the presence of one or more superpoints within a part, each superpoint can query the corresponding part. Unlike bipartite matching, we adopt many-to-one matching~\cite{lu2023query}. Formally, we use a pairwise matching cost matrix $C$ to evaluate the similarity between the queries and the articulation parts. Using the cost matrix, we assign each query to its corresponding parts.

\begin{equation}
\hat{C}_{im} = \begin{cases} 
C_{im} & \text{if } i\text{-th query} \in m\text{-th part} \\
+\infty & \text{otherwise}
\end{cases}
\end{equation}

Once the matching is completed, we know the class labels of the queries in advance and compute the cross-entropy loss $\mathcal{L}_{\text{cls}}$ for each query.
we compute the segmentation mask loss, which consists of the binary cross-entropy loss $\mathcal{L}_{\text{bce}}$ and the dice loss $\mathcal{L}_{\text{dice}}$ for each matched query and part pair. We compute the BCE score loss $\mathcal{L}_{\text{score}}$ to determine if the IoU of the current query's corresponding part is greater than $50\%$.
Therefore, the overall loss for articulation part segmentation is as follows:

\begin{equation}
 \mathcal{L}_{\text{sem}} = \lambda_{\text{cls}} \mathcal{L}_{\text{cls}} + \lambda_{\text{bce}} \mathcal{L}_{\text{bce}} + \lambda_{\text{dice}} \mathcal{L}_{\text{dice}} + \lambda_{\text{score}}  \mathcal{L}_{\text{score}} 
\end{equation}

In our paper, the values of the loss function weights are set as follows: $\lambda_{\text{cls}} = 1.5$, $\lambda_{\text{bce}} = 1.25$, $\lambda_{\text{dice}} = 1.0$, and $\lambda_{\text{score}} = 1.0$, respectively.

%% file: sections/5_Experiment.tex
\section{Experiment} \label{Sec: experiment}

\subsection{Dataset and Evaluation Metrics}

We validate the articulated object segmentation using the GaPartNet dataset~\cite{geng2023gapartnet}, rendering RGB-D images with annotations in the SAPIEN environment~\cite{xiang2020sapien}. To evaluate cross-category generalizability, we split the $27$ object categories into $17$ seen and $10$ unseen categories, ensuring all $9$ part classes are represented in both. Following the 3D semantic instance segmentation benchmarks in ScanNetV2~\cite{dai2017scannet}, we use average precision (AP) as the performance metric for part segmentation, with AP50 (IoU threshold of $50\%$) assessing both per-part and overall segmentation accuracy.

\begin{table}[t]
\caption{Results of Part Segmentation (Per-part-class AP50 \% $\uparrow$ )}
\centering
\label{table:main_result}
\renewcommand\arraystretch{1.2}
\begin{threeparttable}
\resizebox{.49\textwidth}{!}{
\begin{tabular}
{c|c|cccccccccc}
\toprule 
&& Ln.F.Hl. & Rd.F.Hl. & Hg.Hl. & Hg.Ld.& Sd.Ld.  & Sd.Bn  & Sd.Dw. & Hg.Dr. & Hg.Kb. & Avg.AP50
\\ \hline
\multirow{7}{*}{Seen}   & PG\cite{jiang2020pointgroup} &85.3 &23.5 &82.4 &79.3&86.5 &47.3 &60.3 &90.1 &32.5 &65.2 \\ %\cline{2-13} 
                        & SG\cite{vu2022softgroup} & 55.6 & \textbf{92.4} & 80.3 & 75.1 & 87.6 & 26.9 & 50.1 & 91.5 & 50.3& 67.8 \\ %\cline{2-13} 
                        & AGP\cite{liu2022autogpart} &84.3 &22.5  &85.1 &75.6  &87.5  &60.8   &59.7  &91.4  & 18.3 &65.0 \\ %\cline{2-13} 
                        & GAP\cite{geng2023gapartnet}  &89.5 &53.9 & \textbf{90.1} &83.7 & \textbf{89.5} &56.5 & 64.1 & \textbf{93.1} & 51.4 &74.6 \\ 
                        & SPF\cite{sun2023superpoint} & 53.1&43.5&87.3&	76.7&	80.6&	39.5&	52.7&81.2&	15.4& 58.8\\
                      % & SPF-SP & 73.6&46.9&88.2&	78.1&	81.4&	51.8&	54.6&	84.5&	24.2& 64.8 \\
                        & Ours & \textbf{90.4} & 62.0 & 87.5 & \textbf{91.5} & 87.3 & \textbf{71.9} & \textbf{66.4} & 89.8 & \textbf{54.6} & \textbf{77.9}   \\
                        \hline
\multirow{7}{*}{Unseen} & PG\cite{jiang2020pointgroup}  &30.5 &10.9 &2.7 &27.4&0.0 &43.5 &58.3 &60.5 &3.8 &26.4  \\ 
                        & SG\cite{vu2022softgroup} & 24.3 & 7.1 & 1.8 & 35.6 & 0.0 & 49.2 & 52.9 & 67.5 & 11.4 & 27.8  \\ 
                        & AGP\cite{liu2022autogpart} &43.9 &6.4  &3.6  &36.7 &0.0  &46.2 &63.0 &60.6  &15.7  &30.7  \\ 
                        & GAP\cite{geng2023gapartnet} & 43.5 & 37.1 & 2.8 & 40.2 & 3.9 & 45.4 & 60.2 &63.1 &21.2&35.2   \\
                        & SPF\cite{sun2023superpoint} &11.5  &8.7 &1.5 &33.4 &0.5 & 20.2& 10.6&45.4 &3.0 & 15.0\\
                        % & SPF-SP &15.3 &10.6&1.4 &41.3 & 0.8 &35.3 & 16.9 &50.8 &9.7 &20.2  \\
  & Ours & \textbf{46.9} & \textbf{41.2} & \textbf{5.8} & \textbf{42.5} & \textbf{4.1} & \textbf{51.0} & \textbf{65.6} & \textbf{71.4} & \textbf{25.3} & \textbf{39.3}   \\ 
  \toprule 
  % \hline \hline
\end{tabular}
}

\begin{tablenotes} 
{\scriptsize
    \item  Ln.=Line. F.=Fixed. Rd.=Round. Hg.=Hinge. Hl.=Handle. 
     Sd.=Slider.  Ld.=Lid. \item Bn.=Button. Dw.=Drawer. Dr.=Door. 
    Kb.=Knob.  }
    {\tiny
\item PG=PointGroup\cite{jiang2020pointgroup}.   SG=SoftGroup\cite{vu2022softgroup}. 
 AGP=AutoGPart\cite{liu2022autogpart}. GAP=GAPartNet\cite{geng2023gapartnet}.
SPF=SPFormer\cite{sun2023superpoint} 
}
\end{tablenotes}
\end{threeparttable}
\vspace{-1mm}
\end{table}

\begin{table}[t]
\caption{Results of Ablation Studies (Per-part-class AP50 \% $\uparrow$ )}
\centering
\renewcommand\arraystretch{1.2}
\begin{threeparttable}
\resizebox{.49\textwidth}{!}{
\begin{tabular}
{c|c|cccccccccc}
\toprule 
& Ablation & Ln.F.Hl. & Rd.F.Hl. & Hg.Hl. & Hg.Ld.& Sd.Ld.  & Sd.Bn  & Sd.Dw. & Hg.Dr. & Hg.Kb. & Avg.AP50
\\ \hline
\multirow{5}{*}{Seen}   
                        & \xmark \ SP & 56.1 & 25.6 & 65.2 & 63.7 & 53.5 & 42.3 & 46.8 & 65.1 & 31.2 & 49.9 \\ 
                        &  Para. query & 73.6& 46.3& 86.2&	78.1&	81.4&	51.8&	54.6&	84.5&	24.2& 64.5     \\ 
                        & Proj. query & 85.1 & 54.5 & 85.7 & 70.3 & 82.5 & 65.0 & 59.2 & 87.6 & 45.8 &70.6 \\ 
                        &  Ours & \textbf{90.4} & \textbf{62.0} & \textbf{87.5} & \textbf{91.5} & \textbf{87.3} & \textbf{71.9} & \textbf{66.4} & \textbf{89.8} & \textbf{54.6} & \textbf{77.9}     \\ 
                        \hline
\multirow{5}{*}{Unseen} & 
                        \xmark \ SP & 32.5 & 21.5 & 1.9 & 13.7 & 0.0 & 23.7 & 37.6 & 47.4 & 15.5 & 21.5\\ 
                        &  Para. query &15.3 &10.6&1.4 &41.3 & 0.8 &35.3 & 16.9 &50.8 &9.7 &20.2     \\ 
                        & Proj. query &35.9 & 33.5 & 2.1 & 36.4 &0.3& 39.0 & 55.7 & 60.3 & 16.6 & 31.1\\ 
                        &  Ours &\textbf{46.9} & \textbf{41.2} & \textbf{5.8} & \textbf{42.5} & \textbf{4.1} & \textbf{51.0} & \textbf{65.6} & \textbf{71.4} & \textbf{25.3} & \textbf{39.3}    \\ 
  \toprule 
\end{tabular}
}
\begin{tablenotes}
    \item\xmark \ SP= ablate superpoint, Para.query= parameterized query,  \\ Proj. query= center-to-point projection query.
\end{tablenotes}
\end{threeparttable}
\label{table: ablation}
\vspace{-4mm}
\end{table}

\subsection{Cross-category Part Segmentation}
Table~\ref{table:main_result} presents the quantitative comparisons between our method and previous state-of-the-art methods, including PointGroup~\cite{jiang2020pointgroup}, SoftGroup~\cite{vu2022softgroup}, AutoGPart~\cite{liu2022autogpart}, GAPartNet~\cite{geng2023gapartnet}, and SPFormer~\cite{sun2023superpoint}. Our method surpasses current state-of-the-art approaches in both seen and unseen categories, with superior results in $5$ out of $9$ categories. While it shows a slight advantage in seen categories, achieving an AP50 of $77.9\%$, notably, the performance on the slider button category shows an absolute improvement of $11.1\%$ compared to previous methods. Our method performs better in unseen categories, achieving an AP50 of $39.3\%$ and the best results across all part categories, which highlights its enhanced ability to generalize to novel objects.

\begin{figure*}[t]
    \centering
\includegraphics[width=0.96\linewidth]{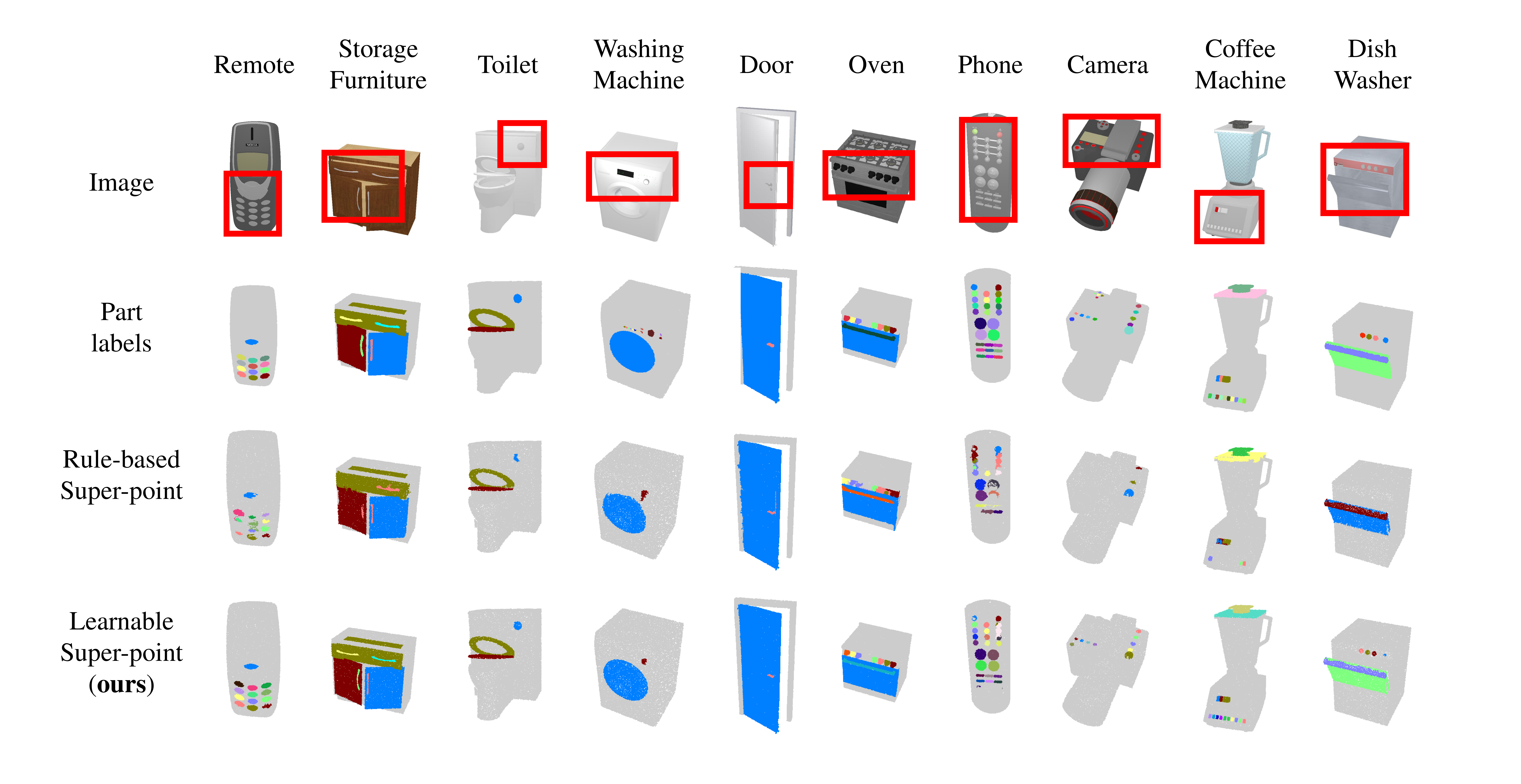}
    \caption{Experimental results of part segmentation using rule-based and learnable superpoints. The segmented parts are marked in red box.}
    \label{Fig: exp results}
\vspace{-3mm}
\end{figure*}

\subsection{Ablation Study}

We conduct ablation studies to validate the effectiveness of the part-aware superpoint generation and SAM-guided 2D information transformer decoder in our method. 

In Table~\ref{table: ablation}, we sequentially ablate the following: superpoint clustering (replaced with raw point clouds), parameterized queries~\cite{sun2023superpoint}, and point-to-center queries~\cite{lu2023query}. Results show that using superpoints instead of raw point clouds significantly improves performance for both seen and unseen objects, as superpoint features capture more transferable geometric information. Compared to parameterized queries~\cite{sun2023superpoint}, the 3D position embeddings of queries better integrate local features, resulting in more accurate part segmentation. Additionally, we modified the query point generation method to a point-to-center query~\cite{lu2023query}, which performed well for seen objects but struggled to accurately locate centers in unseen objects. To address this, SAM leverages 2D prior knowledge to precisely locate part centers through back projection.

We visualize the segmentation results in Figure~\ref{Fig: exp results}, Compared to rule-based superpoints, our learnable superpoint-based queries more effectively integrate local geometric information, enhancing the accuracy of local geometry modeling and improving the overall stability of the model.

%% file: sections/6_Conclusion.tex
\section{Conclusion} \label{Sec: conclusion}

In this paper, we presented Generalizable Articulated Object Perception with Superpoints (GAPS), a novel approach for enhancing part segmentation of articulated objects in 3D point clouds. GAPS employs a learnable, part-aware superpoint generation technique to group points based on geometric and semantic similarities, resulting in clearer boundaries. Furthermore, the method leverages the 2D foundation model SAM to select candidate 3D query points and utilizes a query-based transformer decoder for precise segmentation. GAPS demonstrated state-of-the-art performance on the GAPartNet benchmark, achieving AP50 scores of $77.9\%$ for seen categories and $39.3\%$ for unseen categories, with improvements of $4.4\%$ and $11.6\%$ for seen and unseen categories, highlighting GAPS's generalization capabilities.